\documentclass[11pt,a4paper]{article}
\usepackage{times,latexsym}
\usepackage{url}
\usepackage[T1]{fontenc}

\usepackage[acceptedWithA]{tacl2021v1}
\usepackage{xspace,mfirstuc,tabulary}

\newif\iftaclinstructions
\taclinstructionsfalse % AUTHORS: do NOT set this to true
\iftaclinstructions
\renewcommand{\confidential}{}
\renewcommand{\anonsubtext}{(No author info supplied here, for consistency with
TACL-submission anonymization requirements)}
\newcommand{\instr}
\fi

\iftaclpubformat % this "if" is set by the choice of options

\else

\fi

\usepackage{amsmath}
\usepackage{microtype}
\usepackage[capitalize]{cleveref}
\usepackage{graphicx}
\usepackage{booktabs}
\usepackage{makecell}   
\usepackage[table]{xcolor}

\usepackage{simpleicons}
\newcommand{\hflogo}{\raisebox{-0.1em}{\includegraphics[height=1.2em]{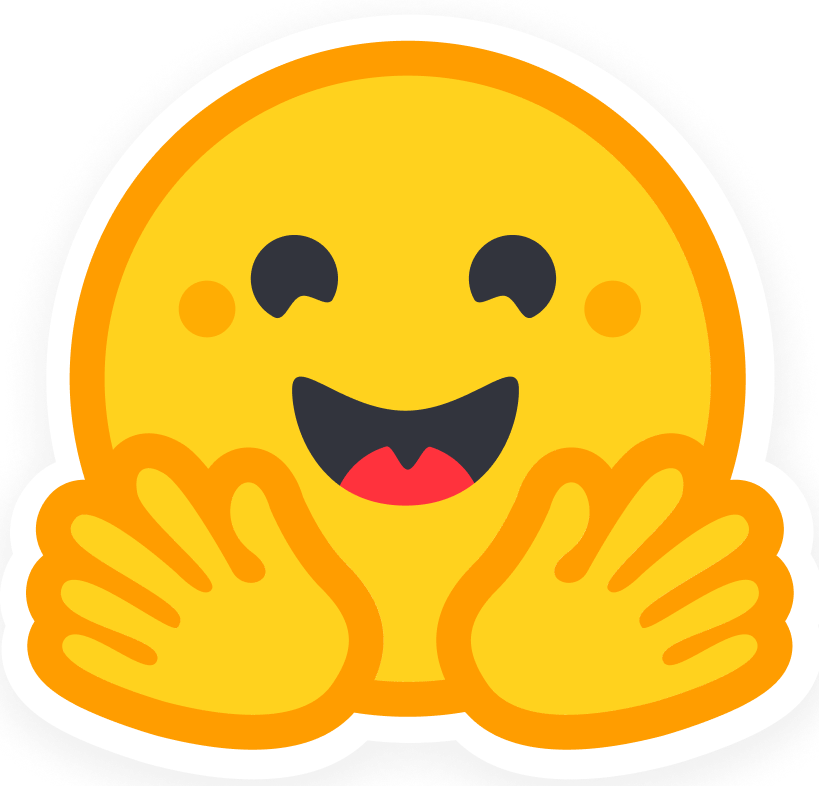}}}
\title{DREAM: Dense Retrieval Embeddings via Autoregressive Modeling}

\author{Yixuan Tang, Yi Yang \\
The Hong Kong University of Science and Technology\\
\texttt{ytangch@connect.ust.hk, imyiyang@ust.hk}\\
  \href{https://github.com/yixuantt/DREAM}{\simpleicon{github}\ GitHub}
  \quad
  \href{https://huggingface.co/collections/yixuantt/dream}{ \hflogo\ Hugging Face}
}

\date{}

\begin{document}
\maketitle

\begin{abstract}
Dense retrieval embedding models are a fundamental component of modern retrieval-based AI systems. Most dense retrievers are trained with contrastive objectives, which require labeled positive and negative document pairs that are often costly and difficult to obtain. In this work, we investigate whether the autoregressive next-token prediction objective of a large language model (LLM) can provide supervision for dense retrieval. The intuition is simple: if a document contains information relevant to a query, conditioning on that document should make the target output easier for the LLM to predict. %Conversely, documents that do not provide useful information should contribute little to prediction. 
A key challenge is that the next-token prediction loss is computed inside the LLM, while the retriever is a separate embedding model. To address this challenge, we propose \textbf{DREAM} (\textbf{D}ense \textbf{R}etrieval \textbf{E}mbeddings via \textbf{A}utoregressive \textbf{M}odeling), which injects retriever-generated query-document similarity scores into selected attention heads of a frozen LLM. During training, these scores determine how much attention each candidate document receives while the LLM predicts the target output. The resulting prediction loss provides gradients for retriever training through the attention mechanism.  We evaluate DREAM on retrieval benchmarks BEIR and RTEB using embedding backbones ranging from 0.5B to 3B parameters. DREAM consistently outperforms existing baselines across different model scales. These results demonstrate that DREAM provides a promising approach for training dense retrievers through autoregressive modeling.

% Dense retrievers are usually trained with contrastive objectives, which need labeled positive documents and carefully mined negatives. These labels are costly, and they say little about whether a document actually helps the model that uses it. We ask whether the next-token prediction loss of a large language model (LLM) can train a retriever instead. The idea is simple: if a document is useful for a query, reading it should make the target output easier to predict. The difficulty is that this loss is computed inside the LLM, so it cannot directly train a separate embedding model. We close this gap by feeding the retriever's query-document scores into a few attention heads of a frozen LLM. The scores decide how much the LLM reads each candidate while predicting the target, so the prediction loss can train the retriever. To lower this loss, the retriever must rank the documents that support the query above the rest of the candidates. The rest serve as negative examples without ever being labeled. On BEIR and RTEB with backbones from 0.5B to 3B, our method outperforms other LLM-supervised retrievers at every scale.  These results show that next-token prediction offers an effective alternative to contrastive supervision, training dense retrievers without labeled positives or negatives.
\end{abstract}

\section{Introduction}

\begin{figure*}[t]
  \centering
  \includegraphics[width=\linewidth]{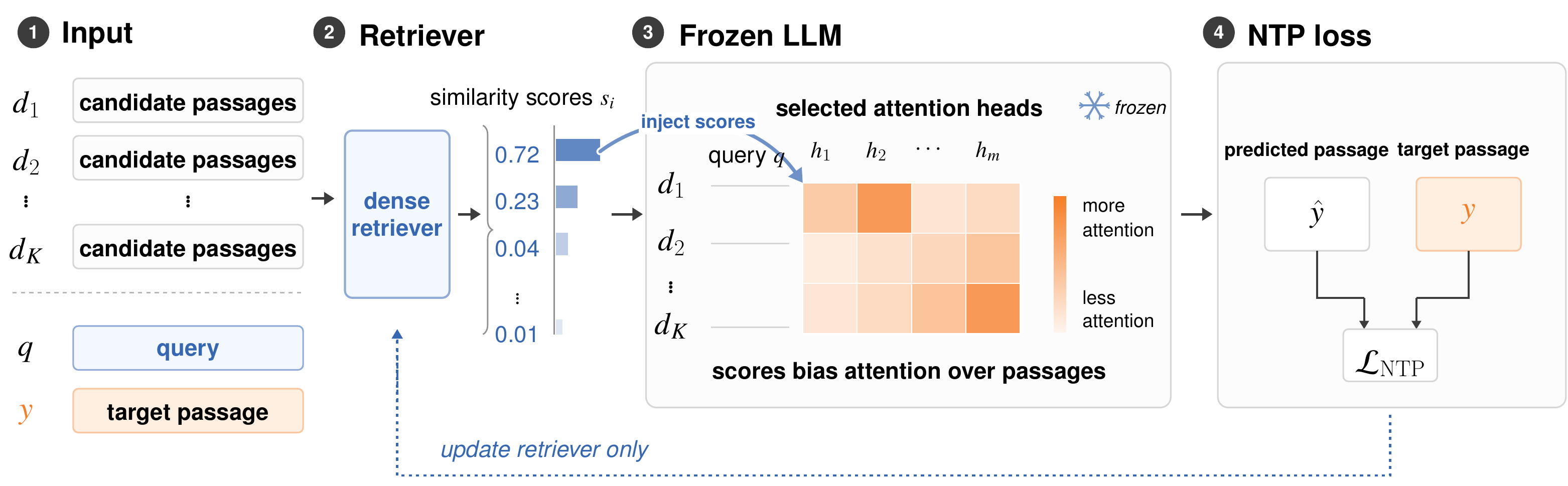}
  \caption{Overview of DREAM. The retriever scores candidate documents for a query, and these scores reweight selected attention heads of a frozen LLM as it predicts the target passage. The next-token prediction loss trains the retriever while the LLM stays frozen.}
  \label{fig:method_overview}
\end{figure*}

Large language model (LLM) systems increasingly rely on retrieval during generation. Retrieval-augmented generation adds external documents to the prompt, and agentic search systems allow an LLM to issue queries, revisit memory, call tools, and plan subsequent actions \citep{lewis2020retrieval,wang2024voyager,ni2026following}.  In these settings, the retriever determines which context is available to the LLM, making retriever training an important part of the overall system.

Most dense retrievers are trained with contrastive objectives. A training example pairs a query with positive documents, which are treated as desired retrieval targets, and sampled negatives, which are treated as lower-relevance alternatives. The objective increases similarity to the positives and decreases similarity to the negatives \citep{reimers2019sentence,gao2021simcse,wang2024improving}. In practice, however, constructing positive and negative examples is often the bottleneck. Positive examples typically require expensive relevance annotations, while hard negatives are difficult to mine reliably and may include false negatives \citep{xiong2021approximate,wang2024mitigating}.%As a result, what the retriever learns depends directly on which documents are chosen as positives and negatives.In practice, constructing those positives and negatives, however, is often the bottleneck. Dense retrievers usually benefit from large training sets, yet many target corpora do not come with existing labels \citep{gao2023precise}. Positives may require human relevance judgments, answer-evidence annotations, or generated queries. Negatives also introduce another risk: random samples often provide little signal, while mined hard negatives can include false negatives that should receive high scores \citep{xiong2021approximate,wang2024mitigating}. %Errors in these choices propagate into the training objective. There is also a deeper issue. These labels fix what counts as relevant before training starts, and they never ask whether a document actually helps the model that will read it. 
%Therefore, this motivates seeking a training signal that does not rely on contrastive labels.
At the same time, autoregressive modeling via next-token prediction (NTP) has become the foundation of modern language-model training \citep{Radford2018ImprovingLU,Radford2019LanguageMA}. Its success suggests that rich supervision can emerge from predicting future tokens in raw text, without any manually specified relevance labels. This observation raises an intriguing question: \textit{can dense retrievers be effectively trained from the autoregressive next-token prediction objective?}

Next-token prediction could potentially provide a useful supervision signal for retrieval. Retrieved documents are ultimately consumed by an LLM to support generation, and their value lies in how much they help the model produce the desired output. If a candidate document contains useful information for answering a query, conditioning on that document should make the target output easier to predict and reduce the next-token prediction loss. Conversely, documents that do not provide useful information should contribute little to prediction and therefore offer little reduction in loss. This gives a direct way to assess retrieval quality: measure its effect on the LLM's prediction loss.

%Consider a simple example. The query asks when a bridge was built, and the target output states the year and the builder. A document that describes the construction of the bridge makes these tokens easy to predict. A document about its daily traffic does not. The prediction loss separates the two documents, and no human label is needed. In other words, next-token prediction replaces labels with a direct test: a document is relevant when it makes the target easier to predict. This test matches how retrieved documents are actually used by the LLM.

However, translating this signal into a training objective for a dense retriever is not straightforward. The autoregressive NTP loss is computed within the LLM, while the retriever is a separate embedding model that ranks documents using query-document similarity scores. If these similarity scores do not influence the LLM’s computation, the loss provides no gradient signal for updating the retriever. The key challenge is therefore to connect the retriever’s similarity scores to the LLM computation during training, while keeping the resulting retriever a standalone embedding model at inference time.

Recent work has begun to explore next-token prediction as a supervision signal for retrieval. RePlug distills document preferences from frozen-LLM likelihoods, but the retriever remains external to the LLM computation \citep{shi2024replug}. Revela {incorporates retriever-computed similarities into language modeling} through in-batch attention, but it trains the language model together with the retriever on raw text batches, without {an explicit query or target output} in the objective \citep{cai2026revela}. As a result, the objective does not directly measure whether a candidate document helps the model produce the desired response. While these methods demonstrate the potential of next-token prediction for retrieval, it remains unclear how to fully exploit autoregressive language modeling as a supervision signal for training dense retrievers.

In this work, we propose DREAM, a method that connects a dense retriever to a frozen LLM through attention, enabling next-token prediction loss to directly supervise retrieval training. The central idea is to use attention as the interface between retrieval and LLM generation: retriever scores determine how much attention is assigned from the query to each candidate document. For each training instance, we concatenate the candidate documents, query, and target output.
The retriever computes a similarity score for each query-document pair, and these scores are injected into selected attention heads of the frozen LLM. As a result, the retriever influences the attention assigned from the query to each candidate document, while the next-token prediction loss provides supervision for updating the retriever. The LLM itself remains frozen throughout training, and the resulting retriever can be used as a standalone embedding model at inference time.

This design is motivated by two considerations. First, not all attention heads are equally relevant for retrieval. Many heads specialize in functions unrelated to query-document matching and therefore provide little information about document relevance. We therefore inject retriever scores only into query-focused retrieval heads identified by \citet{zhang2025queryfocused}. Because these heads already attend from the query to potentially useful context, modulating their attention allows the next-token prediction loss to provide a more informative supervision signal for retrieval.
Second, retriever scores are normalized across candidate documents, creating competition for attention. Because candidates share a fixed attention budget, increasing the weight of one document necessarily reduces the weights of others. This allows the prediction loss to implicitly suppress documents that do not help predict the target output, eliminating the need for explicitly constructed negative examples.

%The design rationale rests on two properties of attention in the frozen LLM. First, injecting the scores into every head would add noise, because most heads compute functions unrelated to retrieval. Therefore, the scores must enter only the heads that perform a retrieval function for the query token. %When the scores enter query-focused retrieval heads, a higher score makes the query read more from a candidate that can help predict the target output, so the loss provides a useful training signal for the ranking. 
%When the same scores enter unrelated heads, they interfere with computations that are not organized around query-document relevance and mostly act as noise; the loss then gives a weak signal about which documents should be preferred. 
%Second, the scores are normalized over the candidate set before they enter attention. Within each selected head, the retriever therefore assigns one shared document-level attention budget across the candidates. Increasing the mass on a useful candidate reduces the mass on the others, so the next-token prediction loss can move attention away from candidates that do not help predict the target without requiring negative labels.

We evaluate the learned retrievers on BEIR \citep{thakur2021beir} and RTEB \citep{muennighoff2023mteb} using embedding backbones ranging from 0.5B to 3B parameters. Across all tested scales, DREAM consistently outperforms RePlug and Revela, with gains ranging from 0.015 to 0.081 NDCG@10 on BEIR and from 0.068 to 0.102 on RTEB. These results demonstrate that next-token prediction can serve as an effective supervision signal for dense retrieval.
{Our analysis further explains why the approach works.} We find that the supervision signal is most effective when retrieval scores are injected into attention heads that already gather evidence for the query. Since the LLM remains frozen, these retrieval heads provide a natural interface through which prediction loss can guide retrieval. In contrast, injecting retrieval scores into randomly chosen heads leads to substantially weaker performance. Overall, the experimental results suggest that autoregressive next-token prediction is a viable alternative to contrastive supervision for training dense retrievers.

\section{Related Work}
This work is related to three areas: {retrieval-model training}, language-model supervision for retrieval, and attention heads in language models.

\paragraph{Retrieval models.}
Most retrieval models learn to map queries and documents into a shared space where a query lies close to its relevant documents. Dense dual-encoders embed a query and a passage separately and rank them by similarity \citep{karpukhin2020dense}, and this contrastive recipe has been strengthened with large-scale labeled data \citep{bajaj2016msmarco} and hard-negative mining \citep{xiong2021approximate,qu2020rocketqa}. Sentence-BERT and SimCSE bring the same matching idea to sentence embeddings \citep{reimers2019sentence,gao2021simcse}, and E5 scales weakly supervised text-pair training \citep{wang2022text}. More recently, decoder-only LLMs serve as embedding backbones and as generators of synthetic query-document pairs when labels are scarce \citep{bonifacio2022inpars,wang2024improving}. Across these architectures, the training signal is still a pairwise matching objective over labeled or synthesized pairs. Constructing such pairs can be costly, and the resulting supervision may be noisy due to annotation errors or false negatives. DREAM takes a different approach by deriving supervision directly from the next-token prediction objective of a frozen LLM rather than from pre-constructed positive and negative pairs.

\paragraph{Language-model supervision for retrieval.}
Another line of work uses a language model to supervise retrieval. REALM jointly trains a retriever with a masked language model so that the language-model objective shapes retrieval \citep{guu2020retrieval}. RePlug trains a retriever from frozen-LLM likelihoods but keeps the retriever scores outside the LLM forward pass \citep{shi2024replug}. Revela trains dense retrievers with a language-modeling objective over chunk sequences, jointly updating the retriever and the language model \citep{cai2026revela}. 
% INTRA retrieves within an encoder-decoder model by using decoder attention queries to score pre-encoded evidence for generation \citep{hoffer2026retrieval}. 
These methods show that language-model feedback can supervise retrieval. {However, RePlug and Revela do not directly train the retriever in the query-candidate-target setting used at inference. RePlug distills preferences from frozen-LLM likelihoods, and Revela optimizes sequential chunk prediction over raw text batches. In both cases, the retriever is not directly supervised by whether a candidate document helps the LLM produce the target output for a given query. DREAM addresses this gap by injecting retriever scores into the frozen LLM computation, so the next-token prediction loss trains the retriever through the documents it weights.}

\paragraph{Attention heads in language models.}
Multi-head attention lets different heads specialize in distinct token-to-token computations \citep{vaswani2017attention}. Interpretability work shows that induction heads implement structured copying that underlies in-context learning \citep{olsson2022incontext}, and more recent work identifies query-focused retrieval heads whose attention links query tokens to the relevant parts of a long context \citep{zhang2025queryfocused}. Prior work mainly uses retrieval heads to analyze how LLMs route information. DREAM instead uses these heads for dense retriever training.

\section{Method}
Each training instance contains a query \(q\), candidate documents \(d_1,\ldots,d_K\), and a target passage \(y\). DREAM trains a dense retriever by feeding the retriever's query-document scores into selected attention heads of a frozen decoder-only LLM as it predicts the target passage. This frozen LLM acts as a judge that evaluates how well the retriever’s selected documents help predict the target output. \cref{fig:method_overview} summarizes the training architecture.

\subsection{Query-Document Similarity Scores}
\label{subsec:scoring}

The trainable model is a dense retriever \(f_\phi\). Given the query and each candidate document, the retriever produces L2-normalized last-token representations:
\begin{equation}
  e_q = \frac{f_\phi(q)}{\|f_\phi(q)\|_2}, \qquad
  e_{d_j} = \frac{f_\phi(d_j)}{\|f_\phi(d_j)\|_2}.
\end{equation}
The query-document similarity score is
\begin{equation}
  s_\phi(q,d_j)=e_q^\top e_{d_j}.
\end{equation}
This is the score used by the final retriever at inference time. During training, we normalize the candidate scores into document-level weights:
\begin{equation}
  p_\phi(d_j \mid q)
  =
  \frac{\exp(s_\phi(q,d_j)/\tau)}
       {\sum_{k=1}^{K}\exp(s_\phi(q,d_k)/\tau)},
\end{equation}
where \(\tau\) is a learnable temperature. The distribution \(p_\phi(d_j\mid q)\) is the document-level signal that enters the frozen LLM during training.

\subsection{Selecting Query-Focused Heads}
\label{subsec:head_selection}
We select the attention heads before training, using the query-focused retrieval-head procedure of \citet{zhang2025queryfocused}. {The goal is to find heads that already perform a retrieval function: their query-token attention assigns higher weight to candidate documents that support the target. We inject retriever scores into these heads because their attention is already organized around query-document relevance. This makes the loss sensitive to whether the retriever gives higher scores to useful documents. Injecting scores into unrelated heads would instead perturb computations that are not about retrieval and provide a noise training signal.}

Each probe example follows the same query-candidate-target format as training. The candidate chunk that supports the target passage is treated as the relevant document for head selection. We place the candidate documents before the query, as in the training input. Let \(D_j\) be the token span of candidate document \(d_j\), and let \(Q(q)\) be the query-token positions. For each head \(h\), we measure how much attention flows from the query tokens to each candidate document:
\begin{equation}
  r_h(q,d_j)
  =
  \frac{1}{|Q(q)|}
  \sum_{a\in Q(q)}
  \sum_{b\in D_j}
  A_h^q(a,b),
\end{equation}
where \(A_h^q(a,b)\) is the post-softmax attention weight in the frozen LLM. This raw score can reflect position or prompt-format bias, not only query-specific evidence use. We therefore compute a query-independent baseline by replacing the query with a content-free query \(q_{\mathrm{null}}\), such as \texttt{N/A}, and subtract it:
\begin{equation}
  \tilde r_h(q,d_j)
  =
  r_h(q,d_j)-r_h(q_{\mathrm{null}},d_j).
\end{equation}
For each head, we rank the candidate documents by \(\tilde r_h(q,d_j)\), compute NDCG@10 against the known relevant documents, and average over the probe set. We keep the top \(M\) heads and denote them by \(H\). Only these heads receive the retriever-guided attention in the next step.

\subsection{Injecting Scores into Attention}
\label{subsec:inject}

During training, the candidate documents, query, and target passage are concatenated into a single decoder-only LLM input. We use document-first causal order:
\begin{equation}
\begin{aligned}
  \mathcal{I} =
  (&\textsc{Passages:}\ d_1,\ldots,d_K\\
   &\textsc{Question:}\ q\\
   &\textsc{Target passage:}\ y).
\end{aligned}
\end{equation}
The candidate documents come first because the query and target tokens can only attend to earlier positions. We use the same \(D_j\) and \(Q(q)\) notation as above. This order lets the query read the candidate documents and lets the target passage read the query. We therefore inject scores only into query-token rows. The scores change which candidate documents the query gathers evidence from, and the target loss evaluates that evidence when predicting \(y\).

For a selected head \(h\in H\) and query-token row \(a\in Q(q)\), DREAM separates attention into two choices: which document to read and which tokens to read inside that document. The retriever should control the first choice, while the frozen LLM should keep the second choice. Let \(\alpha_h(a,b)\) be the original attention from \(a\) to token \(b\). To keep only the LLM's token preference inside document \(d_j\), we normalize the attention over the tokens in \(D_j\):
\begin{equation}
  \hat{\alpha}_{h,j}(a,b)
  =
  \frac{\alpha_h(a,b)}
       {\sum_{b'\in D_j}\alpha_h(a,b')},
  \qquad b\in D_j .
\end{equation}
This normalized distribution sums to one inside \(d_j\). We then multiply it by the retriever weight \(p_\phi(d_j\mid q)\), so the document receives the total mass chosen by the retriever and distributes that mass across its tokens according to the frozen LLM. For \(b\in D_j\),
\begin{equation}
  \alpha^R_h(a,b)
  =
  p_\phi(d_j\mid q)\,\hat{\alpha}_{h,j}(a,b).
\end{equation}
For tokens outside the candidate document spans, \(\alpha^R_h(a,b)=0\).
Thus, \(p_\phi(d_j\mid q)\) controls attention across documents, and \(\hat{\alpha}_{h,j}(a,b)\) controls attention within each document.

Finally, the selected head mixes the original attention with this score-guided attention:
\begin{equation}
  \alpha'_h(a,b)
  =
  (1-g)\,\alpha_h(a,b) + g\,\alpha^R_h(a,b),
\end{equation}
where \(g=\sigma(\gamma)\in[0,1]\) is a learnable gate. We apply this mixture only to selected heads and query-token rows, all other attention rows keep the original attention. The gate controls how strongly the retriever's document choice changes the frozen head.

% The retriever decides how much attention each document receives on the selected heads, the frozen LLM still decides which tokens matter within each document, and the gate controls how strongly the retriever's choice replaces the original attention. The retrieval signal is added without erasing what the head originally does.

\subsection{Training Objective}
\label{subsec:train}
The modified attention is used only during training. The frozen LLM predicts the target passage, and the loss is standard next-token cross entropy on the target-passage tokens.
\begin{equation}
  \mathcal{L}_{\mathrm{NTP}}
  =
  -\sum_{t\in \mathcal{T}_{Y}}
  \log p_\theta(x_t \mid x_{<t}; \{\alpha'_h\}_{h\in H}),
\end{equation}
where \(\theta\) denotes frozen LLM parameters and \(\mathcal{T}_{Y}\) indexes the target-passage tokens. The gradients do not update \(\theta\). They flow from the target-passage loss through \(\alpha'_h\), into \(p_\phi(d_j\mid q)\), through the similarity scores \(s_\phi(q,d_j)\), and finally into the retriever. If a candidate document helps the frozen LLM predict the target passage, increasing its similarity score can reduce the loss. If it does not help, increasing its score is penalized by the same loss.

This objective differs from likelihood distillation in how the supervision reaches the retriever. The next-token prediction loss directly updates the retriever through the modified attention, rather than first producing judge scores for the retriever to imitate. As a result, a score change is useful only when it helps the frozen LLM predict the target passage. The objective also creates competition among candidates without mined negatives. Because \(p_\phi(\cdot\mid q)\) is normalized over the candidate set, increasing the weight of one document lowers the weights of the others. Gradients from the prediction loss therefore move attention toward candidates that help predict the target and away from candidates that do not.

\section{Experiments}
We evaluate whether DREAM can train a stronger standalone retriever. The main experiment compares retrieval performance against lexical and LLM-supervised retrieval baselines, and \cref{sec:analysis} then examines why the next-token prediction signal works and which design choices it depends on.

\subsection{Experimental Setup}
\paragraph{Training data.} We build the training data from the Wikipedia corpus\footnote{\url{https://huggingface.co/datasets/Tevatron/wikipedia-nq-corpus}}. Each document is split into 16 chunks, forming one candidate chunk set. For each set, we choose one chunk as the target passage and ask Qwen3-14B \citep{yang2025qwen3} to generate a query whose answer is supported by that chunk. The target chunk is used as the positive retrieval target, while the full chunk set provides the candidate documents seen during training. The query-generation prompt is provided in \cref{app:query_prompt}.

\paragraph{Implementation.}
DREAM uses a frozen Llama-3.1-8B-Instruct \citep{grattafiori2024llama} as the next-token prediction judge. Unless otherwise stated, we use the top 16 heads from the query-focused retrieval-head ranking, listed in \cref{app:selected_heads}. We train LoRA adapters on the embedding model's \(q/k/v/o\) projection modules with rank 32 and alpha 64. Training uses learning rate \(10^{-4}\), gradient accumulation 32, batch size 1, 1500 steps, and 16 candidate documents per sample.

\paragraph{Baselines.}
The main baselines are BM25, RePlug, and Revela \citep{shi2024replug,cai2026revela}. BM25 is a lexical retrieval baseline and does not use a learned embedding backbone. {InfoNCE~\citep{oord2018representation} is a contrastive baseline trained on the same data and candidate pool as DREAM, using the target chunk as the positive and the other candidates in the same set as negatives. To keep the comparison controlled, we do not apply hard-negative mining, so InfoNCE and DREAM differ only in the training objective.} RePlug is a frozen-LLM likelihood distillation baseline, where the retriever learns from document preferences induced by LLM likelihoods. Revela is an LLM-supervised dense retrieval baseline that trains from language-modeling signals over text chunks. RePlug, Revela, and DREAM are compared at 0.5B, 1B, and 3B embedding scales. The three scales use Qwen2.5-0.5B \citep{team2024qwen}, Llama-3.2-1B, and Llama-3.2-3B \citep{grattafiori2024llama} as the embedding backbones.
% Because InfoNCE training uses a different supervision form, we report it only as an additional result in \cref{app:infonce}.

\paragraph{Benchmarks.}
We evaluate with NDCG@10 on BEIR \citep{thakur2021beir} and RTEB \citep{muennighoff2023mteb}. BEIR covers nine retrieval tasks spanning argument, biomedical, financial, scientific, and community-question-answering domains. RTEB covers fourteen retrieval tasks spanning legal, financial, code, structured-data, and medical domains. Detailed per-task scores are provided in \cref{app:per_task}.
% Task details are provided in \cref{app:benchmark_details}.
% , and full per-task scores are provided in \cref{app:per_task}. 

\begin{table*}[htbp]
\centering
\small
\setlength{\tabcolsep}{6pt}
\renewcommand{\arraystretch}{1.08}
\begin{tabular}{@{}l ccc ccc@{}}
\toprule
 & \multicolumn{3}{c}{\textbf{BEIR}} & \multicolumn{3}{c}{\textbf{RTEB}} \\
\cmidrule(lr){2-4}\cmidrule(lr){5-7}
Method
 & \makecell{Qwen2.5\\0.5B} & \makecell{Llama-3.2\\1B} & \makecell{Llama-3.2\\3B}
 & \makecell{Qwen2.5\\0.5B} & \makecell{Llama-3.2\\1B} & \makecell{Llama-3.2\\3B} \\
\midrule
\textsc{BM25} & \multicolumn{3}{c}{0.4122} & \multicolumn{3}{c}{0.3176} \\
\midrule
InfoNCE & 0.2993 & 0.3268 & 0.3339 & 0.2950 & 0.3658 & 0.3405 \\
\midrule
\textsc{RePlug} & 0.2593 & 0.2535 & 0.2705 & 0.2782 & 0.2855 & 0.3250 \\
\textsc{Revela} & 0.4011 & 0.4075 & 0.4315 & 0.4107 & 0.4499 & 0.4945 \\
DREAM & \textbf{0.4163} & \textbf{0.4888} & \textbf{0.5074} & \textbf{0.4788} & \textbf{0.5514} & \textbf{0.5892} \\
\bottomrule
\end{tabular}
\caption{Average NDCG@10 on BEIR and RTEB. Each benchmark block reports three embedding backbones (Qwen2.5-0.5B, Llama-3.2-1B, Llama-3.2-3B). {BM25 is a lexical baseline with no embedding backbone, so its score is shown once per benchmark.}}
\label{tab:main_results}
\end{table*}

% \begin{table*}[htbp]
% \centering
% \small
% \setlength{\tabcolsep}{7pt}
% \renewcommand{\arraystretch}{1.08}
% \begin{minipage}[t]{0.48\textwidth}
% \centering
% \begin{tabular}{@{}lccc@{}}
% \toprule
% \multicolumn{4}{c}{\textbf{BEIR}} \\
% \midrule
% Method & 0.5B & 1B & 3B \\
% \midrule
% \textsc{BM25} & \multicolumn{3}{c}{0.4122} \\
% \midrule
% InfoNCE & 0.2993 & 0.3268 & 0.3339 \\
% \midrule
% \textsc{RePlug} & 0.2593 & 0.2535 & 0.2705 \\
% \textsc{Revela} & 0.4011 & 0.4075 & 0.4315 \\
% DREAM & \textbf{0.4163} & \textbf{0.4888} & \textbf{0.5074} \\
% \bottomrule
% \end{tabular}
% \end{minipage}
% \hfill
% \begin{minipage}[t]{0.48\textwidth}
% \centering
% \begin{tabular}{@{}lccc@{}}
% \toprule
% \multicolumn{4}{c}{\textbf{RTEB}} \\
% \midrule
% Method & 0.5B & 1B & 3B \\
% \midrule
% \textsc{BM25} & \multicolumn{3}{c}{0.3176} \\
% \midrule
% InfoNCE & 0.2950 & 0.3658 & 0.3405 \\
% \midrule
% \textsc{RePlug} & 0.2782 & 0.2855 & 0.3250 \\
% \textsc{Revela} & 0.4107 & 0.4499 & 0.4945 \\
% DREAM & \textbf{0.4788} & \textbf{0.5514} & \textbf{0.5892} \\
% \bottomrule
% \end{tabular}
% \end{minipage}
% \caption{Average NDCG@10 on BEIR and RTEB. For \refine{InfoNCE,} RePlug, Revela, and DREAM, the columns correspond to Qwen2.5-0.5B, Llama-3.2-1B, and Llama-3.2-3B embedding backbones. BM25 has no embedding backbone, so its score is shown once per benchmark.}
% \label{tab:main_results}
% \end{table*}

\subsection{Main Retrieval Results}

\paragraph{Best average retrieval performance.} \cref{tab:main_results} tests whether the next-token prediction signal produces a stronger retriever under the same embedding backbone. For BM25, DREAM is slightly better on BEIR at 0.5B and clearly better at larger scales, and it is better on RTEB at all scales. DREAM also surpasses InfoNCE at every scale. Since InfoNCE and DREAM use the same data and candidate pool, this gap points to the training objective rather than the shared candidate set. DREAM further outperforms RePlug and Revela, with gains over Revela ranging from 0.015 to 0.081 NDCG@10 on BEIR and from 0.068 to 0.102 on RTEB.

The detailed results in \cref{tab:beir_per_task_main,tab:rteb_per_task_main} show that these gains are not driven by a single dataset. On BEIR, DREAM becomes strongest on nearly all tasks as the backbone grows, including scientific, biomedical, and community-question-answering tasks. On RTEB, the improvements are also broad, with especially large gains on code and structured-data retrieval tasks such as Apps, MBPP, and WikiSQL. This suggests that connecting retrieval scores to selected LLM attention heads gives a stronger and transferable training signal.

\section{Analysis: Why Does the Signal Work?}
\label{sec:analysis}

The main results show that the training signal works. This section asks why. 

\subsection{The Interface Determines the Signal}
\label{subsec:interface_analysis}

\begin{figure}[t]
  \centering
  \includegraphics[width=.95\linewidth]{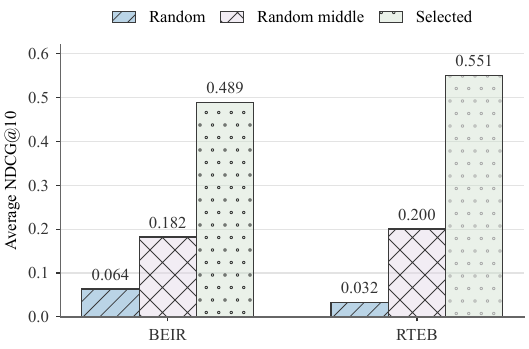}
  \caption{Head selection analysis on BEIR and RTEB with the Llama-3.2-1B backbone. Heads selected from the frozen LLM's query-focused retrieval-head ranking provide much stronger supervision than random heads or random middle-layer heads.}
  \label{fig:head_selection_ablation}
\end{figure}

\paragraph{Effect of head selection.}
The head-selection analysis tests whether retrieval scores can be inserted into any attention head, or whether they must enter heads whose original attention already follows query-candidate relevance. \Cref{fig:head_selection_ablation} shows that the interface matters. Fully random heads reach only 0.0637 average BEIR NDCG@10 and 0.0320 average RTEB NDCG@10, while selected heads reach 0.4888 and 0.5514. Random middle-layer heads improve over fully random heads, but remain far below the selected heads. {This follows from how training works: because the LLM is frozen, DREAM cannot make arbitrary heads learn a new retrieval role.} The retrieval scores are useful only when they modulate heads that already affect how the query reads candidate context. In those heads, increasing the weight of a useful candidate can lower the next-token prediction loss, {so the loss guides the retriever on which candidates to rank higher.} In unrelated heads, the same score changes disturb computations that are not organized around query-candidate relevance, so the resulting gradients are weak or noisy for retrieval.

\begin{figure}[t]
  \centering
  \includegraphics[width=.95\linewidth]{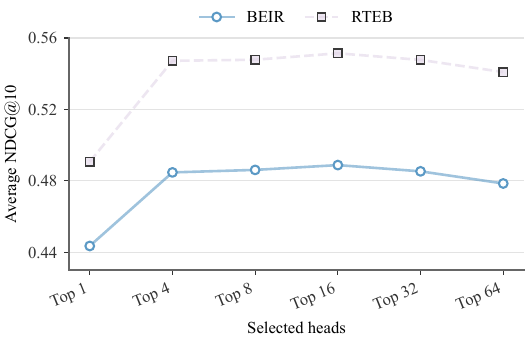}
  \caption{Average NDCG@10 when varying the number of selected heads from the query-focused retrieval-head ranking.}
  \label{fig:head_count_ablation}
\end{figure}

\paragraph{Number of selected heads.}
\Cref{fig:head_count_ablation} shows a clear trend under the same Llama-3.2-1B setting. Performance improves as the number of selected heads increases from Top 1 to Top 16, then drops when the set expands to Top 32 and Top 64. This suggests that one head is not enough to carry the training signal, while adding too many heads can include weaker retrieval heads and dilute the signal. Top 16 provides the best balance in this experiment.

% \begin{table}[htbp]
% \centering
% \small
% \setlength{\tabcolsep}{9pt}
% \renewcommand{\arraystretch}{1.06}
% \begin{tabular}{lcc}
% \toprule
% Selected heads & BEIR & RTEB \\
% \midrule
% Top 1 & 0.4435 & 0.4907 \\
% Top 4 & 0.4847 & 0.5472 \\
% Top 8 & 0.4861 & 0.5478 \\
% Top 16 & \textbf{0.4888} & \textbf{0.5514} \\
% Top 32 & 0.4853 & 0.5477 \\
% Top 64 & 0.4785 & 0.5409 \\
% \bottomrule
% \end{tabular}
% \caption{Average NDCG@10 when varying the number of selected heads from the query-focused retrieval-head ranking.}
% \label{tab:head_analysis}
% \end{table}

\subsection{What the Retriever Learns}
\label{subsec:geometry_analysis}

\begin{figure}[t]
  \centering
  \includegraphics[width=.88\linewidth]{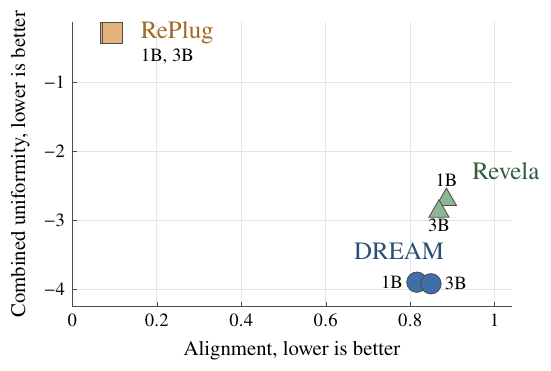}
  \caption{Embedding-space analysis on 5,000 query-positive pairs. Lower is better on both axes. }
  \label{fig:embedding_geometry}
\end{figure}

\paragraph{Embedding geometry.}
We use two embedding-space metrics to understand what the retriever learns beyond retrieval scores. Alignment measures the average squared distance between a query embedding and its positive-document embedding. Uniformity measures how spread out the query and positive-document embeddings are in the representation space~\citep{gao2021simcse}. Lower alignment means positives are closer to their queries, while lower uniformity means the representations are less collapsed. We compute these metrics on 5,000 query-positive pairs randomly sampled from the MTEB retrieval data.

\Cref{fig:embedding_geometry} reports the results for the 1B and 3B embedding backbones. RePlug places positive documents very close to their queries, but the embeddings are poorly spread out. Revela improves the spread of the space, but with weaker alignment. DREAM achieves the best uniformity while keeping alignment in the same range as Revela. The gain is not just better alignment. The retriever also learns a less collapsed embedding space. This better uniformity likely comes from competition among candidates during training. The document weights sum to one over each candidate set, so the retriever is rewarded for pushing unhelpful documents away from the query, not only for pulling the helpful one closer. This embedding geometry is particularly encouraging because DREAM is not trained with a contrastive objective, which is typically associated with improving alignment and uniformity. Despite relying solely on autoregressive next-token prediction, DREAM learns representations with similar geometric properties.

\section{Training Ablations}

\begin{figure}[t]
  \centering
  \includegraphics[width=.88\linewidth]{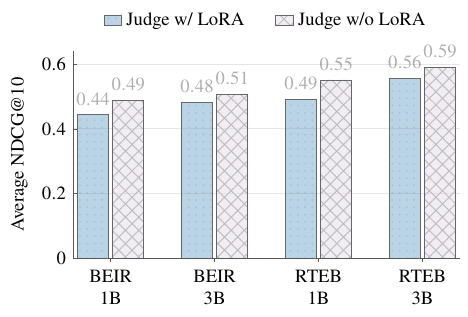}
  \caption{Effect of updating the judge LLM. The frozen-judge setting trains only the embedding model adapters, while the judge LoRA variant also updates LoRA adapters in the LLM.}
  \label{fig:llm_update_ablation}
\end{figure}

\paragraph{Effect of updating the judge LLM.}
\Cref{fig:llm_update_ablation} compares the main frozen-LLM setting with a variant that also adds LoRA adapters to the judge LLM during training. Keeping the judge frozen is better in all four comparisons. This suggests that the fixed LLM computation provides a more stable training signal for the retriever, while updating the judge can weaken the signal that reaches the retrieval model. This result follows from the role of the judge in our objective. {The frozen LLM acts as a fixed judge of document usefulness:} if the retriever gives more weight to documents that help predict the target, the loss should fall. When the LLM is also updated, the loss can fall for another reason: the LLM changes its own prediction behavior. Then a lower loss no longer points as directly to better document weights, so the gradient gives the retriever a weaker training signal. This is why freezing the LLM is useful in our setting. The model that judges document usefulness stays fixed, and only the retriever learns to change which documents it lets the LLM read.

% \begin{table}[t]
% \centering
% \small
% \setlength{\tabcolsep}{5pt}
% \begin{tabular}{lcc}
% \toprule
% Benchmark & LLM-side LoRA & Frozen LLM \\
% \midrule
% BEIR 1B & 0.4435 & \textbf{0.4888} \\
% BEIR 3B & 0.4824 & \textbf{0.5074} \\
% RTEB 1B & 0.4907 & \textbf{0.5514} \\
% RTEB 3B & 0.5575 & \textbf{0.5892} \\
% \bottomrule
% \end{tabular}
% \caption{LLM update ablation. The main setting keeps the LLM frozen and trains only the embedding model adapters.}
% \label{tab:llm_update_ablation}
% \end{table}

\begin{figure}[t]
  \centering
  \includegraphics[width=.92\linewidth]{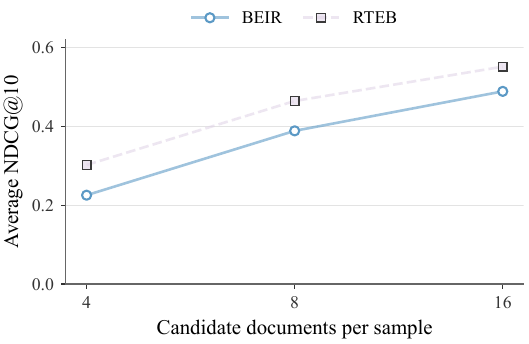}
  \caption{Effect of the number of candidate documents per training sample with the Llama-3.2-1B embedding backbone.}
  \label{fig:candidate_doc_ablation}
\end{figure}

\paragraph{Number of candidate documents.}
\Cref{fig:candidate_doc_ablation} varies the number of candidate documents in each training sample using the Llama-3.2-1B embedding backbone. Increasing the candidate set from 4 to 16 improves average NDCG@10 on both BEIR and RTEB, which suggests that the training loss benefits from having enough alternatives to compare. This fits the competition view of the objective: a larger candidate set gives each training step more documents to compare, which sharpens the signal. We therefore use 16 candidate documents as the default setting.

\paragraph{Scaling to larger models.}
We test whether the same training recipe remains useful with an 8B-scale embedding backbone. Specifically, we train DREAM-8B from Llama-3.1-8B~\citep{grattafiori2024llama} using the same recipe.
\Cref{tab:larger_backbone} compares DREAM-8B with several strong off-the-shelf embedding models, including BGE-large-en-v1.5~\citep{xiao2024c}, Cohere-embed-english-v3.0\footnote{\url{https://cohere.com/blog/introducing-embed-v3}}, E5-mistral-7b-instruct~\citep{wang2024improving}, and Qwen3-Embedding-8B~\citep{zhang2025qwen3}. DREAM-8B reaches 0.5531 average NDCG@10 on BEIR and 0.6417 on RTEB, matching E5-mistral-7b-instruct on BEIR and achieving a 0.0386 higher score on RTEB. While DREAM-8B does not surpass Qwen3-Embedding-8B, this comparison should be interpreted with caution, as Qwen3-Embedding-8B is trained on a different backbone model with a more carefully curated datasets. Our goal in this experiment is not to establish a new state of the art, but to evaluate whether autoregressive next-token prediction remains an effective supervision signal at larger model scales. The strong performance of DREAM-8B suggests that the proposed training paradigm scales favorably and continues to produce competitive retrieval models.

\begin{table}[t]
\centering
\small
\setlength{\tabcolsep}{6pt}
\begin{tabular}{lcc}
\toprule
Model & BEIR & RTEB \\
\midrule
BGE-large-en-v1.5 & 0.5360 & 0.4943 \\
Cohere-embed-english-v3.0 & 0.5406 & 0.5083 \\
E5-mistral-7b-instruct & 0.5526 & 0.6031 \\
\rowcolor{gray!12}
\textbf{DREAM-8B (ours)} & 0.5531 & 0.6417 \\
Qwen3-Embedding-8B & 0.6348 & 0.7383 \\
\bottomrule
\end{tabular}
\caption{Average NDCG@10 with an 8B-scale embedding model. The comparison is included as a scaling check.}
\label{tab:larger_backbone}
\end{table}

\section{Conclusion}
We introduced DREAM, a method for training standalone dense retrievers through autoregressive next-token prediction. By injecting query-document scores into selected attention heads of a frozen LLM, DREAM enables retrieval training without manually labeled relevance pairs. Across BEIR and RTEB, DREAM consistently outperforms existing LLM-supervised retrieval baselines. Our analyses further show that this supervision signal works best through query-focused retrieval heads and naturally produces a better embedding space. Overall, our results suggest that autoregressive next-token prediction is a practical and scalable alternative to contrastive supervision for training retrievers.

\bibliography{tacl2021}
\bibliographystyle{acl_natbib}

\onecolumn

\appendix
\section{Query-Generation Prompt}
\label{app:query_prompt}

We use the following prompt to generate a query from the target passage in each candidate chunk set.
\begingroup
\scriptsize
\begin{verbatim}
Read the following passage and write ONE question that can ONLY be answered from this passage.

Requirements:
- The answer must come from information in this passage.
- Do NOT ask about general knowledge outside the passage.
- The question should require reading more than one sentence to answer.

Output ONLY the question inside the XML tags:

<question>
Your question here
</question>

Passage:
{target_passage}

Question:
\end{verbatim}
\endgroup

\section{Selected Attention Heads}
\label{app:selected_heads}
The default DREAM setting uses the top 16 attention heads from the query-focused retrieval-head ranking of the frozen Llama-3.1-8B-Instruct judge. The ranking is computed on 5,000 examples from the training data before retriever training. \Cref{tab:selected_heads} lists the selected heads. Layer and head indices are zero-based, following the indexing used in the detection code.

\begin{table}[htbp]
\centering
\small
\setlength{\tabcolsep}{10pt}
\renewcommand{\arraystretch}{1.08}
\begin{tabular}{cc@{\hspace{1.2cm}}cc}
\toprule
Rank & Head \((\ell,h)\) & Rank & Head \((\ell,h)\) \\
\midrule
1 & (13, 18) & 9 & (16, 19) \\
2 & (14, 22) & 10 & (17, 26) \\
3 & (14, 13) & 11 & (16, 1) \\
4 & (14, 31) & 12 & (16, 8) \\
5 & (14, 20) & 13 & (20, 1) \\
6 & (13, 1) & 14 & (24, 27) \\
7 & (13, 13) & 15 & (16, 25) \\
8 & (17, 24) & 16 & (13, 21) \\
\bottomrule
\end{tabular}
\caption{Top 16 selected attention heads used by default in DREAM. Each head is shown as (layer, head).}
\label{tab:selected_heads}
\end{table}

\section{Per-Task Retrieval Results}
\label{app:per_task}
\Cref{tab:beir_per_task_main,tab:rteb_per_task_main} report per-task NDCG@10 for the main experiment. These tables supplement the average scores in \cref{tab:main_results}.

\begin{table*}[t]
\centering
\scriptsize
\setlength{\tabcolsep}{3.4pt}
\resizebox{\textwidth}{!}{%
\begin{tabular}{l>{\columncolor{gray!12}}c*{9}{c}}
\toprule
Method & Avg. & ArguAna & NFCorpus & FiQA & SciFact & SCIDOCS & Quora & TREC-COVID & Touche & CQADupStack \\
\midrule
\textsc{BM25} & 0.4122 & 0.4629 & 0.3098 & 0.2339 & 0.6644 & 0.1503 & 0.8067 & 0.6504 & 0.2493 & 0.1818 \\
\addlinespace[3pt]
\multicolumn{11}{l}{\textit{Base model: Qwen2.5-0.5B}} \\
\midrule
\textsc{InfoNCE} & 0.2993 & 0.3670 & 0.1027 & 0.1748 & 0.4590 & 0.0473 & 0.8483 & 0.3455 & 0.1186 & 0.2307 \\
\textsc{RePlug} & 0.2593 & 0.3350 & 0.0543 & 0.1617 & 0.3164 & 0.0111 & 0.8083 & 0.4191 & 0.0891 & 0.1389 \\
\textsc{Revela} & 0.4011 & 0.4262 & 0.2365 & \textbf{0.2830} & \textbf{0.6434} & \textbf{0.1467} & 0.8394 & 0.4808 & \textbf{0.1930} & 0.3610 \\
Ours & \textbf{0.4163} & \textbf{0.5637} & \textbf{0.2522} & 0.2756 & 0.6343 & 0.1376 & \textbf{0.8617} & \textbf{0.5233} & 0.1017 & \textbf{0.3964} \\
\addlinespace[3pt]
\multicolumn{11}{l}{\textit{Base model: Llama-3.2-1B}} \\
\midrule
\textsc{InfoNCE} & 0.3268 & 0.4416 & 0.1033 & 0.2044 & 0.5659 & 0.0662 & 0.8586 & 0.3860 & 0.0642 & 0.2512 \\
\textsc{RePlug} & 0.2535 & 0.3081 & 0.0450 & 0.1776 & 0.3138 & 0.0113 & 0.8221 & 0.4068 & 0.0686 & 0.1287 \\
\textsc{Revela} & 0.4075 & 0.4543 & 0.2587 & 0.3208 & 0.7018 & 0.1726 & 0.8314 & 0.4258 & \textbf{0.1733} & 0.3287 \\
Ours & \textbf{0.4888} & \textbf{0.5762} & \textbf{0.3532} & \textbf{0.3907} & \textbf{0.7243} & \textbf{0.1958} & \textbf{0.8691} & \textbf{0.6983} & 0.1572 & \textbf{0.4343} \\
\addlinespace[3pt]
\multicolumn{11}{l}{\textit{Base model: Llama-3.2-3B}} \\
\midrule
\textsc{InfoNCE} & 0.3339 & 0.4462 & 0.1475 & 0.1912 & 0.5984 & 0.0435 & 0.8449 & 0.4158 & 0.0562 & 0.2613 \\
\textsc{RePlug} & 0.2705 & 0.3726 & 0.0700 & 0.1776 & 0.3718 & 0.0131 & 0.8278 & 0.3711 & 0.0581 & 0.1724 \\
\textsc{Revela} & 0.4315 & 0.4794 & 0.3146 & 0.3503 & 0.7192 & 0.1796 & 0.8298 & 0.4860 & 0.1449 & 0.3800 \\
Ours & \textbf{0.5074} & \textbf{0.5765} & \textbf{0.3806} & \textbf{0.4538} & \textbf{0.7565} & \textbf{0.2135} & \textbf{0.8692} & \textbf{0.6734} & \textbf{0.1614} & \textbf{0.4815} \\
\bottomrule
\end{tabular}%
}
\caption{BEIR per-task NDCG@10. Results are grouped by the base embedding model. BM25 has no learned backbone.}
\label{tab:beir_per_task_main}
\end{table*}
\begin{table*}[t]
\centering
\scriptsize
\setlength{\tabcolsep}{2.2pt}
\resizebox{\textwidth}{!}{%
\begin{tabular}{l>{\columncolor{gray!12}}c*{14}{c}}
\toprule
Method & Avg. & AILA-C & AILA-S & LegalSum & FinBench & HC3Fin & FinQA & Apps & DS1000 & HumanEval & MBPP & WikiSQL & FreshStack & ChatDoctor & CUREv1 \\
\midrule
\textsc{BM25} & 0.3176 & 0.2932 & 0.1646 & 0.5543 & 0.3160 & 0.2655 & 0.7653 & 0.0104 & 0.3297 & 0.3497 & 0.0919 & 0.4365 & 0.2627 & 0.2749 & 0.3325 \\
\addlinespace[3pt]
\multicolumn{16}{l}{\textit{Base model: Qwen2.5-0.5B}} \\
\midrule
\textsc{InfoNCE} & 0.2950 & 0.1407 & 0.1386 & 0.5679 & 0.3106 & 0.2753 & 0.2820 & 0.0317 & 0.2226 & 0.5653 & 0.5465 & 0.4108 & 0.0793 & 0.2024 & 0.3560 \\
\textsc{RePlug} & 0.2782 & 0.1530 & 0.1759 & 0.3285 & 0.3360 & 0.2139 & 0.4280 & 0.0238 & 0.3571 & 0.4186 & 0.4614 & 0.3038 & 0.1232 & 0.2026 & 0.3691 \\
\textsc{Revela} & 0.4107 & \textbf{0.2975} & 0.1595 & \textbf{0.5780} & 0.3054 & 0.3752 & \textbf{0.4579} & 0.1054 & 0.4585 & 0.7950 & 0.7489 & 0.5327 & 0.1956 & 0.3433 & \textbf{0.3969} \\
Ours & \textbf{0.4788} & 0.1957 & \textbf{0.2428} & 0.5145 & \textbf{0.4639} & \textbf{0.3812} & 0.3974 & \textbf{0.2656} & \textbf{0.5523} & \textbf{0.9005} & \textbf{0.8170} & \textbf{0.9176} & \textbf{0.2445} & \textbf{0.4942} & 0.3167 \\
\addlinespace[3pt]
\multicolumn{16}{l}{\textit{Base model: Llama-3.2-1B}} \\
\midrule
\textsc{InfoNCE} & 0.3658 & 0.1618 & 0.2048 & 0.5737 & 0.4821 & 0.2968 & 0.4076 & 0.0593 & 0.2570 & 0.6722 & 0.6006 & 0.5876 & 0.1444 & 0.3019 & 0.3719 \\
\textsc{RePlug} & 0.2855 & 0.1460 & 0.1829 & 0.4046 & 0.4700 & 0.2326 & 0.4722 & 0.0197 & 0.2749 & 0.3052 & 0.2797 & 0.4192 & 0.1565 & 0.2356 & 0.3975 \\
\textsc{Revela} & 0.4499 & \textbf{0.2764} & 0.2062 & \textbf{0.6037} & 0.5071 & 0.4201 & \textbf{0.5456} & 0.1587 & 0.5447 & 0.7661 & 0.7137 & 0.5897 & 0.2477 & 0.3380 & 0.3809 \\
Ours & \textbf{0.5514} & 0.2420 & \textbf{0.3078} & 0.5945 & \textbf{0.7095} & \textbf{0.5402} & 0.4874 & \textbf{0.2530} & \textbf{0.5888} & \textbf{0.8962} & \textbf{0.8298} & \textbf{0.9451} & \textbf{0.2931} & \textbf{0.5649} & \textbf{0.4679} \\
\addlinespace[3pt]
\multicolumn{16}{l}{\textit{Base model: Llama-3.2-3B}} \\
\midrule
\textsc{InfoNCE} & 0.3405 & 0.1863 & 0.1809 & 0.5010 & 0.4794 & 0.3543 & 0.3698 & 0.0272 & 0.2607 & 0.6148 & 0.4141 & 0.5432 & 0.1102 & 0.3248 & 0.3999 \\
\textsc{RePlug} & 0.3250 & 0.1622 & 0.1370 & 0.4858 & 0.5541 & 0.2599 & 0.4538 & 0.0354 & 0.3232 & 0.6053 & 0.3020 & 0.3666 & 0.1632 & 0.2741 & 0.4279 \\
\textsc{Revela} & 0.4945 & 0.2606 & 0.2529 & \textbf{0.6272} & 0.6105 & 0.4878 & \textbf{0.5369} & 0.2055 & 0.5632 & 0.8655 & 0.7698 & 0.6196 & 0.2872 & 0.4093 & 0.4275 \\
Ours & \textbf{0.5892} & \textbf{0.2983} & \textbf{0.3196} & 0.6254 & \textbf{0.7985} & \textbf{0.6457} & 0.4768 & \textbf{0.3860} & \textbf{0.6034} & \textbf{0.9523} & \textbf{0.8708} & \textbf{0.8401} & \textbf{0.3320} & \textbf{0.5998} & \textbf{0.5005} \\
\bottomrule
\end{tabular}%
}
\caption{RTEB per-task NDCG@10. Results are grouped by the base embedding model. BM25 has no learned backbone.}
\label{tab:rteb_per_task_main}
\end{table*}

\end{document}